\documentclass[nohyperref]{article}

\usepackage{microtype}
\usepackage{graphicx}
\usepackage{subfigure}
\usepackage{booktabs} 

\usepackage{hyperref}

\usepackage[accepted]{icml2025_nofootnote}

\usepackage{amsmath}
\usepackage{amssymb}
\usepackage{mathtools}
\usepackage{amsthm}

\usepackage{multirow}
\usepackage{pgfplots}

\usepackage[capitalize,noabbrev]{cleveref}

\newcommand{\bea}{\begin{eqnarray}}
\newcommand{\eea}{\end{eqnarray}}
\newcommand{\be}{\begin{equation}}
\newcommand{\ee}{\end{equation}}

\theoremstyle{plain}

\theoremstyle{definition}

\theoremstyle{remark}

\usepackage[textsize=tiny]{todonotes}

\icmltitlerunning{ Scaling Bidirectional Spans and Span Violations in Attention Mechanism }

\begin{document}

\twocolumn[
\icmltitle{ Scaling Bidirectional Spans and Span Violations in Attention Mechanism 
           }




\begin{icmlauthorlist}
\icmlauthor{Jongwook Kim}{yyy,comp}
\icmlauthor{Sangheon Yun}{yyy,comp}
\icmlauthor{Sukjin Yoon}{snu}
\end{icmlauthorlist}

\icmlaffiliation{yyy}{Department of Physics, Sogang University, Seoul, Korea}
\icmlaffiliation{comp}{IndigoWave, Seoul, Korea}
\icmlaffiliation{snu}{Department of Physics Education, Seoul National University, Seoul, Korea}

\icmlcorrespondingauthor{Jongwook Kim}{dr.jongwookkim@gmail.com}
\icmlcorrespondingauthor{Sangheon Yun}{ sangheon.yun@gmail.com}

\icmlkeywords{Machine Learning, ICML}

\vskip 0.3in
]



\printAffiliationsAndNotice{}  

\begin{abstract}
The canonical $O(N^2)$ Transformer remains the empirical performance frontier in sequence modeling, and its training can be further optimized by addressing geometric inefficiency. We propose an optimization framework that leverages an asymmetric projection to decompose the backward-pass gradients into parallel spans and orthogonal violations, while keeping the canonical forward-pass $QKV$ structure intact. Through consistent experimental validation across various decomposition and projection setups, we provide strong theoretical evidence: the standard attention gradient is suboptimal. We demonstrated that selectively scaling these components, focusing primarily on $0^{th}$ order bidirectional parallel spans, yields the most effective learning signal. On the limited WikiText-2 dataset, and using a crude configuration, this method achieved a $0.56\%$ reduction in validation loss, confirming the framework's fundamental validity and suggesting significant potential gains on larger datasets and deeper training regimes

\end{abstract}

\section{Introduction}
\label{introduction}
While canonical attention mechanisms \cite{vaswani2017attention} remain the most powerful for $O(N^2)$ complexity models, they carry all information throughout the $QKV$ matrices during computation, leading to inefficient processing of irrelevant components. 
We propose a geometrically motivated refinement that achieves superior training efficiency and model performance \emph{without modifying the standard $QKV$ attention mechanism}. We regard the canonical attention Eq.~\ref{eq:attn} as a fixed structural prior and introduce our method as an orthogonal enhancement. Specifically, we decompose $QKV$ into their respective \emph{spans} and \emph{span violations}, and apply gradient scaling to constrain attention to emphasize semantically relevant components while suppressing noise \cite{kedia2024transformersstableendtoendsignal} \cite{Ramaswamy2023icpram23} \cite{zhang2020gradientclippingacceleratestraining}. A key observation is that not all orthogonal components (span violations) are harmful, nor are all parallel components (spans) necessarily beneficial. Building on this, we deliberately focus on \emph{bidirectional} span and span violation as a principled means of isolating informative structure from noise.

\section{Parameter Decomposition}
The standard canonical attention mechanism  \cite{vaswani2017attention} is given by
\be
\mathrm{Attn} = \mathrm{softmax}\left( \frac{Q K^\top}{\sqrt{d}} \right) V, 
\label{eq:attn}
\ee 
where $Q = X W^{Q}$, $K = X W^{K}$, and $V = X W^{V}$, with $X$ of the input sequence length $T$ times feature dimension $d$ and $W^Q, W^K, W^V \in \mathbb{R}^{d\times d}$ learned linear transformations. Our method preserves this forward pass architecture entirely.

When decomposing the $QKV$ matrices, three projection strategies can be considered: left-acting, right-acting, and combined left-right operations. 

Projections of spans and span violations are defined by $\Pi^\parallel + \Pi^\perp = I$. For brevity, we discard the superscript and use $\Pi$ for the projection of spans to denote $\Pi^\parallel$. We adopt the left-acting approach:
\begin{align}
Q &= (\Pi_K + \Pi_K^\perp)(\Pi_V + \Pi_V^\perp) Q, \nonumber \\
K &= (\Pi_Q + \Pi_Q^\perp)(\Pi_V + \Pi_V^\perp) K, \nonumber \\
V &= (\Pi_Q + \Pi_Q^\perp)(\Pi_K + \Pi_K^\perp) V, \nonumber 
\end{align}
where the projection operators are defined as
\begin{align}
\Pi_K &= K (K^\top K)^{-1} K^\top, \quad \nonumber \\
\Pi_V &= V (V^\top V)^{-1} V^\top, \quad \nonumber \\
\Pi_Q &= Q (Q^\top Q)^{-1} Q^\top. \nonumber
\end{align}
Each $\Pi$ is a $(T \times T)$ matrix requiring inversion of a $(d \times d)$ matrix. Right-acting projections, by contrast, require inversion of larger $(T \times T)$ matrices on smaller $(d \times d)$ operators, which is less efficient since $d \ll T$. Combined left-right projections are possible but incur greater computational cost without clear advantages and are therefore omitted.

Even within left-acting projections, there are $2^3 = 8$ different symmetric orderings possible.
Due to the non-commutativity of projection operators ($\Pi_K \Pi_V \neq \Pi_V \Pi_K$), these orderings induce distinct learning dynamics. Asymmetric projections, which apply bidirectional projections only to $Q$ while using unidirectional projections for $K$ and $V$, offer computational efficiency by halving the number of score components to 8 from 16 while maintaining the essential mathematical structure. Moreover, between $(\Pi_V + \Pi_V^{\perp})(\Pi_K + \Pi_K^{\perp})Q$ and $(\Pi_K + \Pi_K^{\perp})(\Pi_V + \Pi_V^{\perp})Q$, we choose the former asymmetric projection as
\bea
Q &=& (\Pi_V + \Pi_V^{\perp})(\Pi_K + \Pi_K^{\perp})Q, \nonumber \\
K &=& (\Pi_V + \Pi_V^{\perp})K, \nonumber \\
V &=& (\Pi_K + \Pi_K^{\perp})V. 
\label{eq:decomposition}
\eea
The ordering $(\Pi_V + \Pi_V^{\perp})(\Pi_K + \Pi_K^{\perp})Q$ applies projections right-to-left: first $\Pi_K$, then $\Pi_V$, producing information flow $Q \to K\text{-space} \to V\text{-space}$ consistent with attention's $Q \to K \to V$ structure. In contrast, the alternative ordering reverses this flow to $Q \to V \to K$, which contradicts the natural direction of attention. We adopt the current asymmetric projection ordering to preserve the semantic and dynamic structure of attention, potentially enabling efficient training.

\section{Score Matrix Decomposition}
\subsection{Unidirectional Decomposition}
To formulate a clean baseline for the full bidirectional setting, key subspace is split into unidirectional way:
\bea
S = \frac{Q \ K^\top}{\sqrt{d}} = \frac{1}{\sqrt{d}} [(\Pi_K + \Pi_K^\perp)]  Q  K^\top 
 \equiv S^{\parallel} + S^{\perp}. \label{eq:unidirectional_decomposition}
\eea

With Einstein index convention where $i,j,a$ range from $1$ to $T$ for sequence positions and $m,n,b$ range from $1$ to $d$ for feature dimensions, the Gram matrix $G = K^\top K \in \mathbb{R}^{d\times d}$ is expressed as $G_{mn} = K_{im} K_{in}$, and the Gram matrix inverse in component form is $G^{mn} = (K^\top K)^{-1}_{mn} \in \mathbb{R}^{d\times d}$. It is convenient to use the Moore-Penrose pseudoinverse:
\be 
K^+ = (K^\top K)^{-1}K^\top \in \mathbb{R}^{d\times T} 
\ee
The projection matrix along $K$ is $(\Pi_K)_{ij} = K_{im} (K^\top K)^{-1}_{mn} K^\top_{nj}$ in component form. 
In accordance with the unidirectional decomposition (\ref{eq:unidirectional_decomposition}), the gradients are calculated as
\begin{alignat}{1}
\frac{\partial L}{\partial Q_{ab}}
  &\!\!=\!\!
   \sum_{ij}^T \sum_{A \in \{\parallel, \perp\}}
   \frac{\partial L}{\partial S_{ij}^A}
   \frac{\partial S_{ij}^A}{\partial Q_{ab}} \nonumber\\
  &\!\!=\!\!
   \sum_{ij}^T \frac{1}{\sqrt{d}}
   \left[
      \frac{\partial L}{\partial S_{ij}^{\parallel}} (\Pi_K)_{ai}
      + \frac{\partial L}{\partial S_{ij}^{\perp}} (\Pi_K^{\perp})_{ai}
   \right] K_{jb}\,,
\end{alignat}

\begin{alignat}{1}
\frac{\partial L}{\partial K_{ab}}
  &\!\!=\!\! 
  \sum_{ij}^T \sum_{A \in \{\parallel, \perp\}}
  \frac{\partial L}{\partial S_{ij}^A}
  \frac{\partial S_{ij}^A}{\partial K_{ab}}
  \nonumber\\[4pt]
  &\!\!=\!\!
  \frac{1}{\sqrt{d}} \sum_{ij}^T
    \Big[
      \Big(
        \frac{\partial L}{\partial S_{ij}^{\parallel}}
        -
        \frac{\partial L}{\partial S_{ij}^{\perp}}
      \Big) \nonumber
      \\[-2pt] &\!\!\!\!\!\!\!\!\times\!\!
      \Big(
        (\Pi_K^{\perp} Q K^\top)_{aj} (K^+)_{bi}
        +
        (\Pi_K^{\perp})_{ai} (K^+ Q K^\top)_{bj}
      \Big)
    \Big]
  \nonumber\\[6pt]
  &\!\!+\!\!
   \frac{1}{\sqrt{d}} \sum_{i}^T
    \left[
      \frac{\partial L}{\partial S_{ia}^{\parallel}}
      (\Pi_K Q)_{ib}
      +
      \frac{\partial L}{\partial S_{ia}^{\perp}}
      (\Pi_K^{\perp} Q)_{ib}
    \right].
\end{alignat}
Q-gradients form a simple parallel/orthogonal split, while K-gradients include additional $QK^\top$ and $K^{+}$ interaction terms.

\subsection{Bidirectional Score Matrix Decomposition}
The left-actions on $Q$ and $K$ induce a corresponding left-right decomposition on the attention score matrix, $S$. We first define the set $\mathcal{A}$ of all possible 4-tuples of projections:
$$\mathcal{A} = \{(\alpha, \beta, \gamma, \delta) | \alpha, \beta, \gamma, \delta \in \{\parallel, \perp\}\}\,,$$
Applying the projection decompositions to $Q$ and $K$, the score matrix $S = Q K^\top / \sqrt{d}$ can be decomposed as:
$$\frac{Q K^\top}{\sqrt{d}} = \frac{1}{\sqrt{d}} \sum_{\alpha,\beta,\gamma,\delta \in  \mathcal{A}} \Pi_{V}^{\alpha} \Pi_{K}^{\beta} Q K^\top \Pi_{K}^{\gamma} \Pi_{V}^{\delta} = \sum_{B=1}^{16} S^B$$
Due to orthogonality constraints, only 8 of the 16 possible component terms are non-zero, allowing the score matrix to be represented as a summation over 8 non-zero basis matrices: $S = \sum_{B=1}^8 S^B$.  \footnote{ $S = \sum_{B=1}^8 S^B +  \sum_{C=9}^{16} S^C$, where  $\sum_{C=9}^{16} S^C$ = 0 since every $S^C = 0$. }

The 8 non-zero score matrix components $S^B$ are categorized based on their order of span violations, defined by the count of orthogonal projection terms $\Pi^\perp \in \{\Pi_K^\perp, \Pi_V^\perp\}$:

\begin{itemize}
\item \textbf{0th order} (Parallel accross $Q$ span and $K$ span): 
\be
S^1 = \frac{1}{\sqrt{d}} \Pi_V \Pi_K Q K^\top \Pi_K \Pi_V \label{eq:score0th}
\ee
\item \textbf{1st order} (Single violation): 
\bea
S^2 &= \frac{1}{\sqrt{d}} \Pi_V \Pi_K Q K^\top \Pi_K \Pi_V^\perp \nonumber \\
S^3 &= \frac{1}{\sqrt{d}} \Pi_V \Pi_K^\perp Q K^\top \Pi_K \Pi_V \nonumber \\
S^5 &= \frac{1}{\sqrt{d}} \Pi_V^\perp \Pi_K Q K^\top \Pi_K \Pi_V \label{eq:score1st}
\eea
\item \textbf{2nd order} (Double violations):
\bea
S^4 &= \frac{1}{\sqrt{d}} \Pi_V \Pi_K^\perp Q K^\top \Pi_K \Pi_V^\perp \nonumber \\
S^6 &= \frac{1}{\sqrt{d}} \Pi_V^\perp \Pi_K Q K^\top \Pi_K \Pi_V^\perp \nonumber \\
S^7 &= \frac{1}{\sqrt{d}} \Pi_V^\perp \Pi_K^\perp Q K^\top \Pi_K \Pi_V \label{eq:score2nd}
\eea
\item \textbf{3rd order} (Triple violations): 
\be
S^8 = \frac{1}{\sqrt{d}} \Pi_V^\perp \Pi_K^\perp Q K^\top \Pi_K \Pi_V^\perp \label{eq:score3rd}
\ee
\end{itemize}

The Frobenius inner product $\langle A,B\rangle = \operatorname{Tr}(A^{\top}B)$ with the Einstein index convention is used to define the total derivative of the loss function as $dL = \operatorname{Tr}\!\big[(\tfrac{\partial L}{\partial S})^{\top} dS\big] = \sum_{ij}^T \frac{\partial L}{\partial S_{ij}}\, dS_{ij}$, where $i,j$ index sequence positions. Most pairs of the $8$-score basis matrices are orthogonal, with exceptions $\langle S^1, S^3\rangle$, $\langle S^2, S^4\rangle$, $\langle S^5, S^7\rangle$, and $\langle S^6, S^8\rangle$. This non-orthogonality arises from the non-commuting projections, $\Pi_V \Pi_K \neq \Pi_K \Pi_V$.

Although the score matrix is block diagonalized in this basis, the decomposition is utilized strictly as an analytical tool for the backward pass gradient analysis, ensuring no modification to the forward pass token sequence ordering.

\section{Gradients}

\subsection{Q Gradient}
The $Q$ derivatives are\footnote{Since $Q$, $K$, and $V$ are independent: $\frac{\partial Q}{\partial K} = \frac{\partial Q}{\partial V} = 0$, $\frac{\partial K}{\partial Q} = \frac{\partial K}{\partial V} = 0$, $\frac{\partial V}{\partial Q} = \frac{\partial V}{\partial K} = 0$.}
\bea
\frac{\partial L}{\partial Q_{ab}} &=& \sum_{B=1}^8 \sum_{ij}^T \frac{\partial L}{\partial S_{ij}^B}   \frac{\partial S_{ij}^B}{\partial Q_{ab}} = \sum_{B=1}^8 \frac{\partial L}{\partial Q_{ab}}^{(B)}
\eea
with index convention where $i,j,a$ denote sequence positions and $b$ ranges from $1$ to $d$ for feature dimension.

The gradient is calculated in order of span violations, \textit{i.e.}, in multiples of orthogonal projections $\Pi^\perp$:
\begin{align}
\frac{\partial L}{\partial Q}_{0\text{th}} &= \frac{\partial L}{\partial Q}^{(1)}
= \frac{1}{\sqrt{d}} (\Pi_V \Pi_K)^\top \frac{\partial L}{\partial S^1} (\Pi_V K), \label{eq:q_grad_0th}
\end{align}

\begin{align}
\frac{\partial L}{\partial Q}_{1\text{st}} &= \frac{\partial L}{\partial Q}^{(2)} + \frac{\partial L}{\partial Q}^{(3)} + \frac{\partial L}{\partial Q}^{(5)} \nonumber\\
&= \frac{1}{\sqrt{d}} (\Pi_V \Pi_K)^\top \frac{\partial L}{\partial S^2} (\Pi_V^{\perp} K) \nonumber\\
& \quad+ \frac{1}{\sqrt{d}} (\Pi_V \Pi_K^{\perp})^\top \frac{\partial L}{\partial S^3} (\Pi_V K) \nonumber\\
& \quad+ \frac{1}{\sqrt{d}} (\Pi_V^{\perp} \Pi_K)^\top \frac{\partial L}{\partial S^5} (\Pi_V K), \label{eq:q_grad_1st}
\end{align}

\begin{align}
\frac{\partial L}{\partial Q}_{2\text{nd}} &= \frac{\partial L}{\partial Q}^{(4)} + \frac{\partial L}{\partial Q}^{(6)} + \frac{\partial L}{\partial Q}^{(7)} \nonumber\\
&= \frac{1}{\sqrt{d}} (\Pi_V \Pi_K^{\perp})^\top \frac{\partial L}{\partial S^4} (\Pi_V^{\perp} K) \nonumber\\
& \quad+ \frac{1}{\sqrt{d}} (\Pi_V^{\perp} \Pi_K)^\top \frac{\partial L}{\partial S^6} (\Pi_V^{\perp} K) \nonumber\\
& \quad+ \frac{1}{\sqrt{d}} (\Pi_V^{\perp} \Pi_K^{\perp})^\top \frac{\partial L}{\partial S^7} (\Pi_V K), \label{eq:q_grad_2nd}
\end{align}

\begin{align}
\frac{\partial L}{\partial Q}_{3\text{rd}} &= \frac{\partial L}{\partial Q}^{(8)} = \frac{1}{\sqrt{d}} (\Pi_V^{\perp} \Pi_K^{\perp})^\top \frac{\partial L}{\partial S^8} (\Pi_V^{\perp} K). \label{eq:q_grad_3rd}
\end{align}
Thus, the total gradient
\be
\frac{\partial L}{\partial Q} = \frac{\partial L}{\partial Q}_{0\text{th}} + \frac{\partial L}{\partial Q}_{1\text{st}} + \frac{\partial L}{\partial Q}_{2\text{nd}} + \frac{\partial L}{\partial Q}_{3\text{rd}}.
\ee

\subsection{K Gradient}
The gradient with respect to $K$ decomposes as

\bea
\frac{\partial L}{\partial K_{ab}} &=& \sum_{B=1}^8 \sum_{ij}^T \frac{\partial L}{\partial S_{ij}^B}   \frac{\partial S_{ij}^B}{\partial K_{ab}}\,.
\eea
We calculate $\frac{\partial \Pi_K}{\partial K}$ as:
\be 
\frac{\partial (\Pi_K)_{ij}}{\partial K_{ab}} = (\Pi_K^{\perp})_{aj}{K^+}_{bi} + (\Pi_K^{\perp})_{ai}(K^+)_{bj} = \frac{\partial (\Pi_K)_{ji}}{\partial K_{ab}}. 
\ee
Utilizing this expression, we calculate the $K$ derivatives in order of span violations:
\be
\frac{\partial L}{\partial K}_{0\text{th}} = \frac{1}{\sqrt{d}} \Pi_V^\top \left(\frac{\partial L}{\partial S^1}\right)^\top \Pi_V \Pi_K Q,
\ee
\bea
\frac{\partial L}{\partial K}_{1\text{st}} &=& \frac{\partial L}{\partial K}_{1\text{st}}^{\text{direct}} + \frac{\partial L}{\partial K}_{1\text{st}}^{\text{cross}}, \\
\frac{\partial L}{\partial K}_{1\text{st}}^{\text{direct}} &=& \frac{1}{\sqrt{d}} \Pi_V^{\perp \top} \left(\frac{\partial L}{\partial S^2}\right)^\top \Pi_V \Pi_K Q \nonumber \\
&+& \frac{1}{\sqrt{d}} \Pi_V^\top \left(\frac{\partial L}{\partial S^3}\right)^\top \Pi_V \Pi_K^{\perp} Q \nonumber \\
&+& \frac{1}{\sqrt{d}} \Pi_V^\top \left(\frac{\partial L}{\partial S^5}\right)^\top \Pi_V^{\perp} \Pi_K Q, \\
\frac{\partial L}{\partial K}_{1\text{st}}^{\text{cross}} &=& \frac{1}{\sqrt{d}} \Pi_K^{\perp} Q K^\top \Pi_V \left(\frac{\partial L}{\partial S^1} - \frac{\partial L}{\partial S^3}\right)^\top \Pi_V K_+^\top \nonumber \\
&+& \frac{1}{\sqrt{d}} \Pi_K^{\perp} \Pi_V \left(\frac{\partial L}{\partial S^1} - \frac{\partial L}{\partial S^3}\right) \Pi_V K Q^\top K_+^\top, \nonumber \\
\eea
\bea
\frac{\partial L}{\partial K}_{2\text{nd}} &=& \frac{\partial L}{\partial K}_{2\text{nd}}^{\text{direct}} + \frac{\partial L}{\partial K}_{2\text{nd}}^{\text{cross}}, \\
\frac{\partial L}{\partial K}_{2\text{nd}}^{\text{direct}} &=& \frac{1}{\sqrt{d}} \Pi_V^{\perp \top} \left(\frac{\partial L}{\partial S^4}\right)^\top \Pi_V \Pi_K^{\perp} Q \nonumber \\
&+& \frac{1}{\sqrt{d}} \Pi_V^{\perp \top} \left(\frac{\partial L}{\partial S^6}\right)^\top \Pi_V^{\perp} \Pi_K Q \nonumber \\
&+& \frac{1}{\sqrt{d}} \Pi_V^\top \left(\frac{\partial L}{\partial S^7}\right)^\top \Pi_V^{\perp} \Pi_K^{\perp} Q, \\
\frac{\partial L}{\partial K}_{2\text{nd}}^{\text{cross}} &=& \frac{1}{\sqrt{d}} \Pi_K^{\perp} Q K^\top \Pi_V^{\perp} \left(\frac{\partial L}{\partial S^2} - \frac{\partial L}{\partial S^4}\right)^\top \Pi_V K_+^\top \nonumber \\
&+& \frac{1}{\sqrt{d}} \Pi_K^{\perp} \Pi_V \left(\frac{\partial L}{\partial S^2} - \frac{\partial L}{\partial S^4}\right) \Pi_V^{\perp} K Q^\top K_+^\top \nonumber \\
&+& \frac{1}{\sqrt{d}} \Pi_K^{\perp} Q K^\top \Pi_V \left(\frac{\partial L}{\partial S^5} - \frac{\partial L}{\partial S^7}\right)^\top \Pi_V^{\perp} K_+^\top \nonumber \\
&+& \frac{1}{\sqrt{d}} \Pi_K^{\perp} \Pi_V^{\perp} \left(\frac{\partial L}{\partial S^5} - \frac{\partial L}{\partial S^7}\right) \Pi_V K Q^\top K_+^\top, \nonumber \\
\eea
\bea
\frac{\partial L}{\partial K}_{3\text{rd}} &=& \frac{\partial L}{\partial K}_{3\text{rd}}^{\text{direct}} + \frac{\partial L}{\partial K}_{3\text{rd}}^{\text{cross}}, \\
\frac{\partial L}{\partial K}_{3\text{rd}}^{\text{direct}} &=& \frac{1}{\sqrt{d}} \Pi_V^{\perp \top} \left(\frac{\partial L}{\partial S^8}\right)^\top \Pi_V^{\perp} \Pi_K^{\perp} Q, \\
\frac{\partial L}{\partial K}_{3\text{rd}}^{\text{cross}} &=& \frac{1}{\sqrt{d}} \Pi_K^{\perp} Q K^\top \Pi_V^{\perp} \left(\frac{\partial L}{\partial S^6} - \frac{\partial L}{\partial S^8}\right)^\top \Pi_V^{\perp} K_+^\top \nonumber \\
&+& \frac{1}{\sqrt{d}} \Pi_K^{\perp} \Pi_V^{\perp} \left(\frac{\partial L}{\partial S^6} - \frac{\partial L}{\partial S^8}\right) \Pi_V^{\perp} K Q^\top K_+^\top. \nonumber \\
\eea
Total $K$ gradient is
\be
\frac{\partial L}{\partial K} = \frac{\partial L}{\partial K}_{0\text{th}} + \frac{\partial L}{\partial K}_{1\text{st}} + \frac{\partial L}{\partial K}_{2\text{nd}} + \frac{\partial L}{\partial K}_{3\text{rd}}.
\ee

\subsection{V Gradient}
The $V$ gradient has the same form as the standard attention, unaffected by the choice of our decomposition Eq.~(\ref{eq:decomposition}). This is due to the fact that $\sum_{B=1}^{8} \frac{\partial S^B_{ij}}{\partial V_{ab}} = \frac{\partial \sum_{B=1}^{8} S^B_{ij}}{\partial V_{ab}} = 0$,
\be
\frac{\partial L}{\partial V} = A^\top \frac{\partial L}{\partial \text{Attn}} \,.
\ee

\section{Gradient Scaling by the Order of Span Violations}
The proposed methodology is a post-backpropagation operation that leaves the forward pass of the standard attention mechanism unchanged. Given the score gradients $\frac{\partial L}{\partial S^B}$ from the backward pass, we apply a set of non-negative valued scale factors $\alpha_0, \alpha_1, \alpha_2, \alpha_3 $ to the decomposed gradient components. The final updated gradients for the query and key weight matrices are expressed as a linear combination of these components, respectively:
\be
\frac{\partial L}{\partial Q} = \sum_{i=0}^3 \alpha_i \frac{ \partial L}{\partial Q}_{i \text{th}} \,\, \text{and}\quad \frac{\partial L}{\partial K} = \sum_{i=0}^3 \alpha_i \frac{ \partial L}{\partial K}_{i \text{th}} .
\ee
Note that the gradients for the Value ($V$) matrix, $\frac{\partial L}{\partial V} $, are not scaled.

\section{Experimental Setup}

We conduct experiments on the WikiText-2 dataset to validate our gradient decomposition methods under two prediction paradigms.

\textbf{Dataset}: WikiText-2 raw version is tokenized using the GPT-2 tokenizer (vocabulary size 50,257). We construct sequences with 50$\%$ overlap for both training and validation sets, yielding language modeling tasks with next-token prediction targets.

\textbf{Model Architecture}: We employ a simple transformer language model with learnable token and position embeddings, followed by multi-head attention layers with 4:1 feedforward expansion ratio, layer normalization, and a final output projection to vocabulary space. We vary the model dimension, number of attention heads, number of transformer layers, and sequence length across experiments, with GELU activation and dropout rate 0.1. 

\textbf{Gradient Accumulation}: With an effective batch size of 128, we observe stable performance across scales.

\textbf{Causal Prediction}: For causal language modeling, we apply a triangular causal mask to the attention mechanism, ensuring each position attends only to previous positions in the sequence. This follows the standard GPT-style autoregressive generation paradigm, where the model predicts the next token given all previous tokens.


\textbf{Baseline Experiments}: We conduct baseline experiments by modulating the gradients of $Q$, $K$, and $V$ matrices using scalar multipliers 
$
\alpha_Q, \alpha_K, \alpha_V \in \{0, 1\} \label{qkv_multiplier}
$
, where $0$ disables and $1$ enables the respective gradient. This establishes a baseline training behavior for comparison with our decomposition methods.

The standard Transformer setting (\texttt{QKV111}) achieves the best performance among all baselines, whereas \texttt{QKV000} performs the worst. Interestingly, \texttt{QKV001} shows strong early-epoch performance but plateaus later, suggesting that gradients through the $V$ matrix dominate early training while $Q$ and $K$ gradients become critical in later stages \cite{michel2019sixteenheads} \cite{rogers2021primerbertology}.
\begin{table}[t]
\centering
\caption{Validation loss minimum with gradient modulation of $Q$, $K$, and $V$: Model architecture $T=512, d=256$, 16 heads($d_{Head}\equiv \frac{d}{H}=16$), 6 layers.  (lower is better)}
\label{tab:qkv_baseline_perf}
\begin{tabular}{lcccc}
\toprule
Setting & $\alpha_Q$ & $\alpha_K$ & $\alpha_V$ & Performance \\
\midrule
 \texttt{QKV111} & 1 & 1 & 1 & 5.5104 \\
 \texttt{QKV110} & 1 & 1 & 0 & - \\
 \texttt{QKV101} & 1 & 0 & 1 & 5.5177 \\
 \texttt{QKV100} & 1 & 0 & 0 & - \\
 \texttt{QKV011} & 0 & 1 & 1 & 5.5286 \\
 \texttt{QKV010} & 0 & 1 & 0 & - \\
 \texttt{QKV001} & 0 & 0 & 1 & 5.5689 \\
 \texttt{QKV000} & 0 & 0 & 0 & 5.5737 \\
\bottomrule
\end{tabular}
\end{table}

\begin{figure}[ht]
\vskip 0.1in
\begin{center}
\includegraphics[width=\columnwidth]{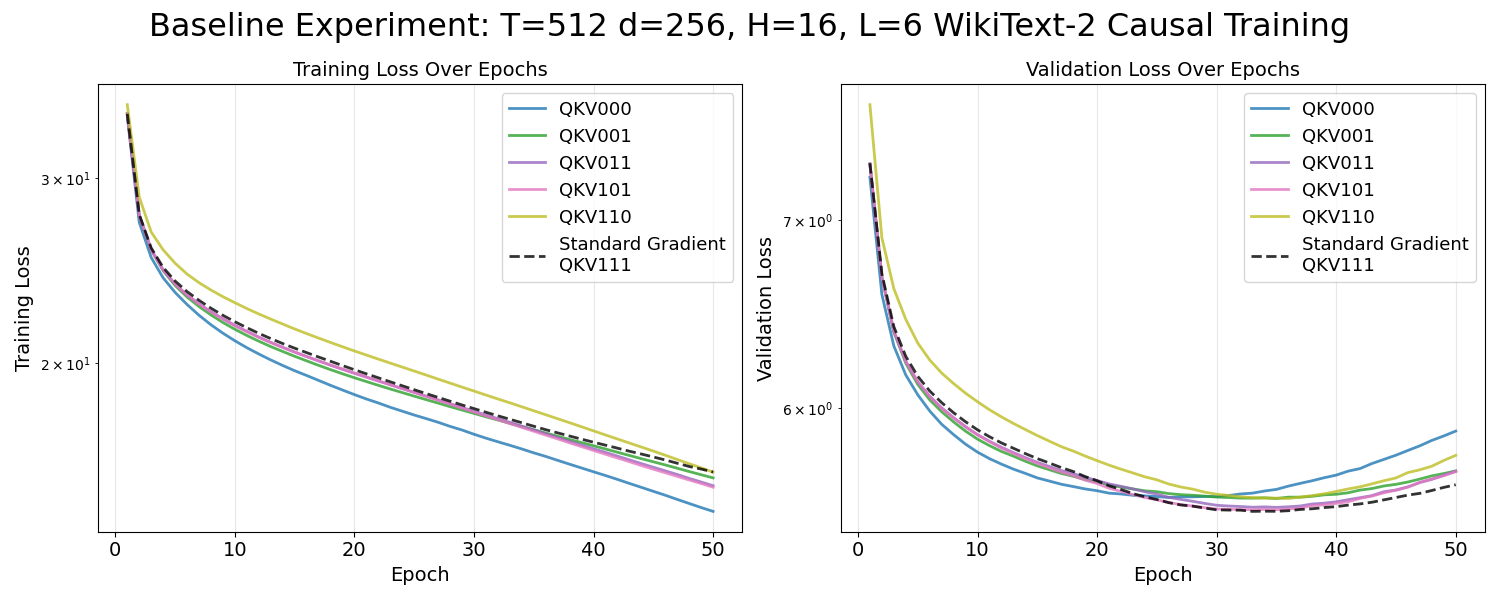}
\vskip 0.1in
\includegraphics[width=\columnwidth]{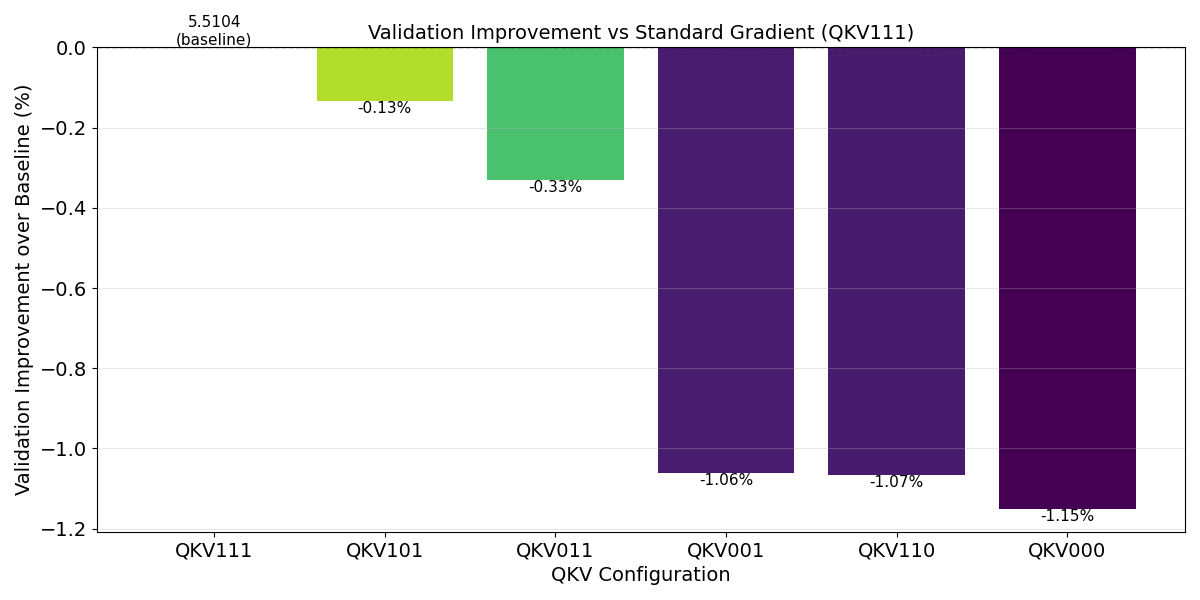}
\caption{ \textbf{Baseline Experiments} - performance comparison for modulating $QKV$ gradients: Model architecture $T=512, d=256$, 16 heads($d_{Head}\equiv \frac{d}{H}=16$), 6 layers.}
\label{fig:baseline_causal_H16L6}
\end{center}
\end{figure}

\section{Reductionistic Gradient Decomposition Experiments}
We evaluate reductionistic gradient decomposition methods that solely rely on the total score gradient $\frac{\partial L}{\partial S}$, thereby circumventing the need for individual score block gradients $\frac{\partial L}{\partial S^B}$. The motivation for these experiments is to verify whether simplified optimization through span decomposition can effectively guide training.
\subsection{Simplest Decomposition Framework}
For the initial analysis, we can first consider a simple gradient decomposition:
\bea
\frac{\partial L}{\partial Q}_{\text{standard}} \!\!\!\!\! &=& \!\!\!\!\! \frac{\partial L}{\partial S} \frac{K}{\sqrt{d}} = (\Pi_K + \Pi_K^\perp) \frac{\partial L}{\partial S} \frac{K}{\sqrt{d}},\\
\frac{\partial L}{\partial K}_{\text{standard}} \!\!\!\!\! &=& \!\!\!\!\! \left(\frac{\partial L}{\partial S}\right)^\top \!\!\! \frac{Q}{\sqrt{d}} = (\Pi_K + \Pi_K^\perp) \left(\frac{\partial L}{\partial S}\right)^\top \!\!\! \frac{Q}{\sqrt{d}}\,. \nonumber
\eea
The scalar multipliers $\alpha_\|$ or $\alpha_\bot$ may adjust gradient components:
\bea
\frac{\partial L}{\partial Q} &=& (\alpha_\|\Pi_K + \alpha_\bot\Pi_K^\perp)\frac{1}{\sqrt{d}} \frac{\partial L}{\partial S} K , \\
\frac{\partial L}{\partial K} &=& (\alpha_\|\Pi_K + \alpha_\bot\Pi_K^\perp) \frac{1}{\sqrt{d}} \left(\frac{\partial L}{\partial S}\right)^\top Q.
\eea

\subsection{Reductionistic Decomposition Framework}

We investigate the approximation of our 4-component decomposition method using $Q$ and $K$ span decompositions, specifically by projecting gradients onto combinations of $\Pi_K$, $\Pi_K^\perp$, $\Pi_V$, and $\Pi_V^\perp$ as: 
\bea
\frac{\partial L}{\partial Q}_{\text{standard}} &=& \frac{1}{\sqrt{d}} \frac{\partial L}{\partial S} K = \sum_{i=0}^3 \frac{\partial L}{\partial Q}_i ,\\
\frac{\partial L}{\partial K}_{\text{standard}} &=& \frac{1}{\sqrt{d}} \left(\frac{\partial L}{\partial S}\right)^\top Q = \sum_{i=0}^3 \frac{\partial L}{\partial K}_i
\eea
, where the decomposition employs projection combinations
\bea
\Pi_0 &=& \Pi_K \Pi_V \Pi_K \nonumber \\
\Pi_1 &=& \Pi_K \Pi_V^\perp \Pi_K \nonumber \\
\Pi_2 &=& \Pi_K^\perp \Pi_V \Pi_K^\perp \nonumber \\
\Pi_3 &=& \Pi_K^\perp \Pi_V^\perp \Pi_K^\perp
\eea
satisfying $\sum_{i=0}^3 \Pi_i = I$. Specifically, $\frac{\partial L}{\partial Q}_i$ applies $\Pi_i$ to $\frac{1}{\sqrt{d}} \frac{\partial L}{\partial S} K$, while $\frac{\partial L}{\partial K}_i$ applies $\Pi_i$ to $\frac{1}{\sqrt{d}} \left(\frac{\partial L}{\partial S}\right)^\top Q$.

To selectively enable or disable gradient components, we apply non-negative scalar multipliers $\alpha_i$ for $i \in \{0,1,2,3\}$:
\be
\frac{\partial L}{\partial Q} = \sum_{i=0}^3 \alpha_i \frac{\partial L}{\partial Q}_i \quad \text{and}\quad \frac{\partial L}{\partial K} = \sum_{i=0}^3 \alpha_i \frac{\partial L}{\partial K}_i.
\ee
\subsection{Check of the Numerical Soundness of Reductionistic Gradient Decomposition} 
\textbf{Standard gradient limit}: We compare the standard $Q, K, V$ gradients with  $\alpha_Q = \alpha_K = \alpha_V = 1$, {\it i.e.} (\texttt{QKV111}) of the baseline experiments and the decomposed gradients with scale factors $\alpha_0 = \alpha_1 = \alpha_2 = \alpha_3 = 1$, {\it i.e.} (\texttt{[1111]}).

\textbf{$V$ gradient-only limit}: The $V$ gradient only training, {\it i.e.} with  $\alpha_Q = \alpha_K = 0$ and $\alpha_V = 1$, {\it i.e.} (\texttt{QKV001}) is compared with the all zero-scale factors experiment {\it i.e.} $\alpha_0 = \alpha_1 = \alpha_2 = \alpha_3 = 0$, {\it i.e.} (\texttt{[0000]}). 
\begin{figure}[ht]
\begin{center}
\includegraphics[width=\columnwidth]{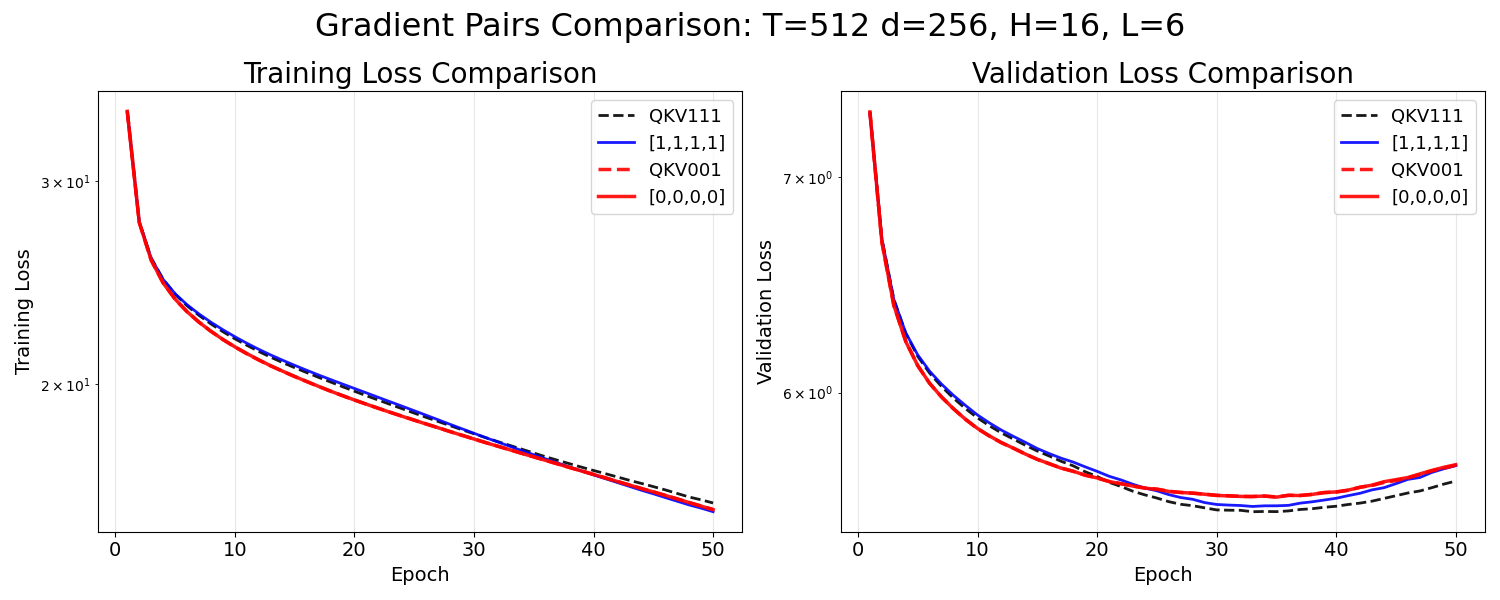}
\caption{\textbf{Base Comparisons}: Model architecture $T=512, d=256$, 4 heads, 6 layers.}
\label{fig:direct_causal_base_pairs_H16L6}
\end{center}
\end{figure}

\section{Score Matrix Decomposition Experiments}

\subsection{8-Component Score Gradient Method}

We implement an 8-component score matrix decomposition using PyTorch autograd hooks to compute block-wise gradients without explicit differentiation of the complete attention mechanism.

\subsubsection{Forward Pass: Score Block Decomposition}

Given query, key, and value matrices $Q, K, V \in \mathbb{R}^{T \times d}$ and their associated projection operators $\{\Pi_K, \Pi_K^\perp, \Pi_V, \Pi_V^\perp\}$, we define 8 projection operator pairs in accordance with (\ref{eq:score0th}), (\ref{eq:score1st}), (\ref{eq:score2nd}), (\ref{eq:score3rd}). The standard attention score matrix $S = \frac{1}{\sqrt{d}} Q K^\top$ is decomposed into 8 score blocks: $S^{B},  B \in \{1, \ldots, 8\}$. 
For causal prediction, each score block is masked with the causal mask before softmax computation. For contextual prediction, each block is left intact. 

\subsubsection{Backward Pass: Autograd Hook for Score Block Gradients}

For each score block $S^{B}$, we compute its gradient $\frac{\partial L}{\partial S^{B}}$ using PyTorch's autograd mechanism:
\begin{enumerate}
\item Detach $S^{B}$ from the computational graph and reattach with \texttt{requires\_grad=True}
\item Compute the attention output: $A^{B} = \text{softmax}(S^{B}) V$
\item Use \texttt{torch.autograd.grad} to compute:
\be
\frac{\partial L}{\partial S^{B}} = \text{autograd}\left(A^{B}, S^{B}, \frac{\partial L}{\partial \text{attn\_output}}\right)
\ee
\end{enumerate}

\subsection{Check of the Numerical Soundness of Score Matrix Gradient Decomposition} 
\begin{figure}[ht]
\vskip 0.1in
\begin{center}
\includegraphics[width=\columnwidth]{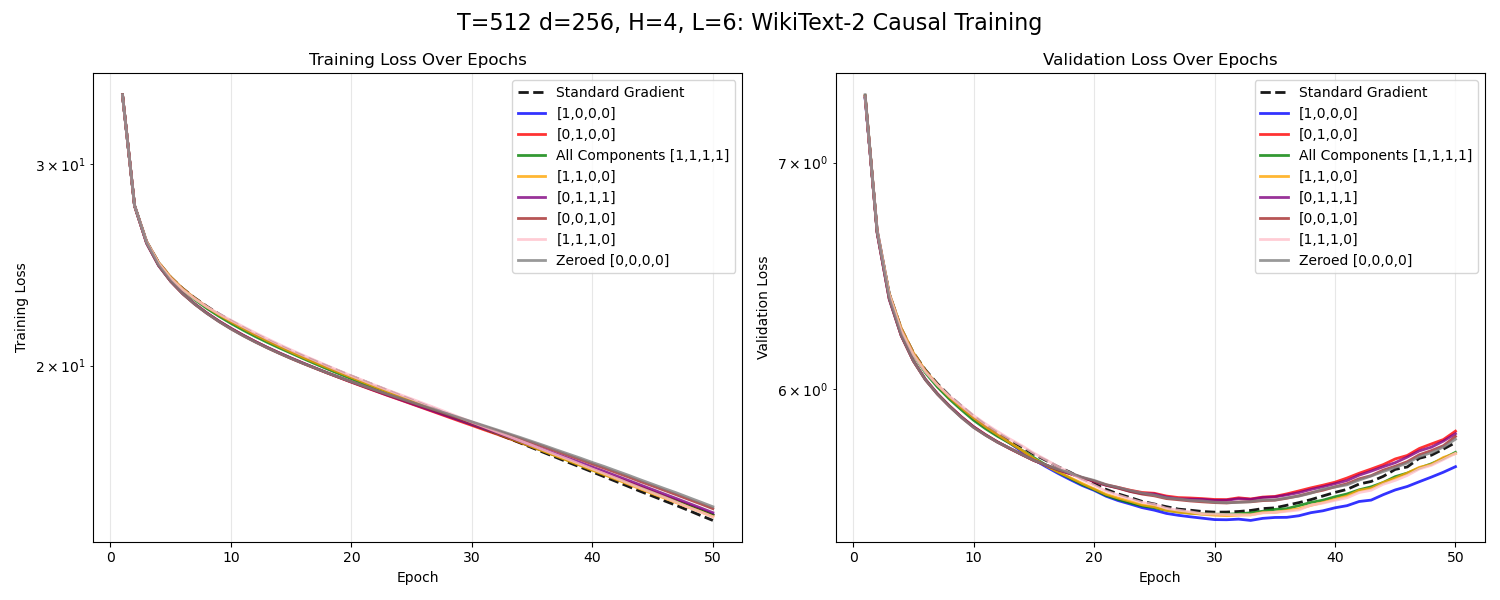}
\vskip 0.1in
\includegraphics[width=\columnwidth]{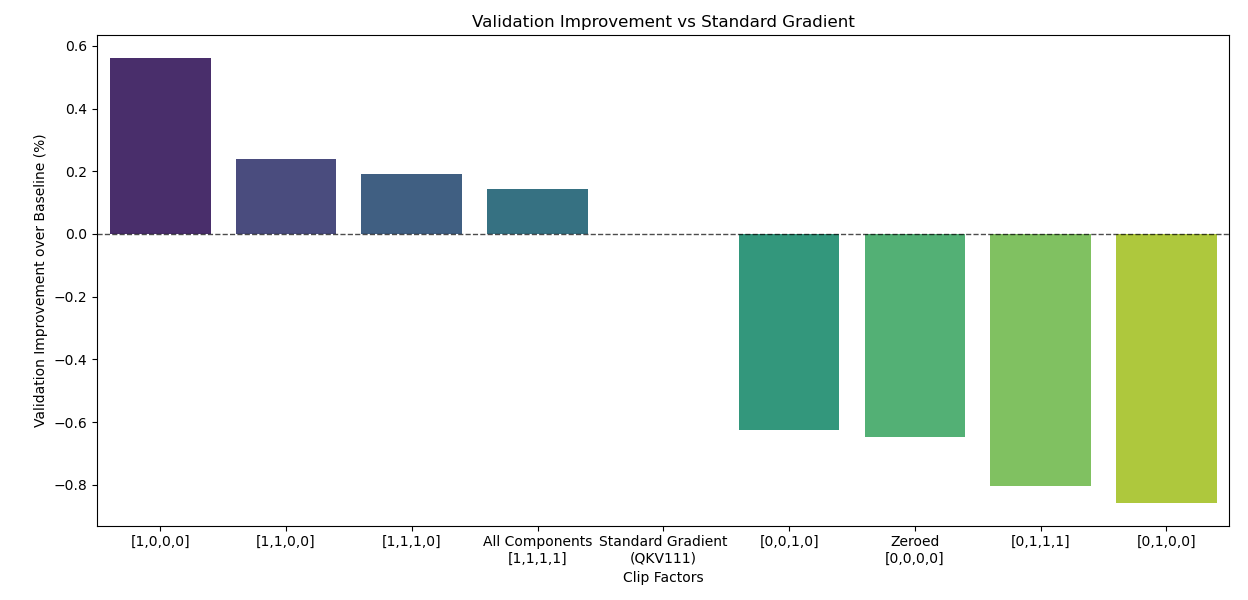}
\caption{\textbf{Score Matrix Decompositions}: Model architecture $T=512, d=256$, 4 heads($d_{Head}\equiv \frac{d}{H}=64$), 6 layers. Top: Training curves show the progression over 50 epochs on the WikiText-2 dataset causal prediction. Dashed line is the Standard Gradient baseline. Bottom: Improvements on the Standard Gradient baseline validation loss minimum. }
\label{fig:hook_causal_H4L6}
\end{center}
\vskip -0.1in
\end{figure}

\textbf{Standard gradient limit}: Using the standard gradient experiment (Standard Gradient) as our performance reference, we first compare it with the nominal full-scaling configuration (\texttt{QKV111}– \texttt{[1111]}) as presented in Figure. \ref{fig:hook_causal_H4L6}. Although \texttt{[1111]} is designed to be mathematically equivalent to the standard gradient, a small discrepancy remains (best-val 5.5089 vs. 5.5167). This mismatch originates from numerical errors introduced when computing the 8-block aggregated gradient $S^B$ through PyTorch’s autograd hook mechanism(\texttt{torch.autograd.grad}).
\begin{figure}[ht]
\begin{center}
\includegraphics[width=\columnwidth]{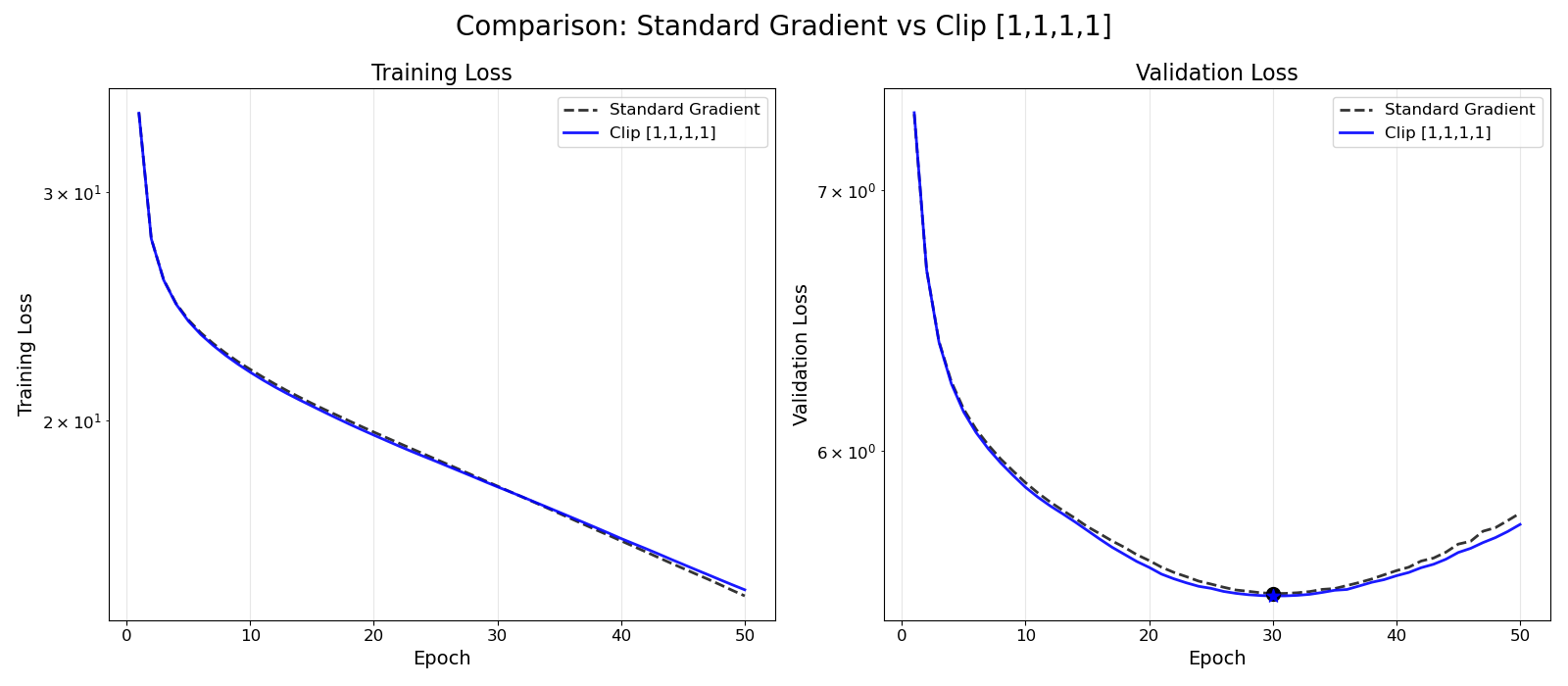}
\caption{\textbf{Base Comparisons}: Model architecture $T=512, d=256$, 4 heads, 6 layers.}
\label{fig:hook_causal_H4L6_qkv111_1111}
\end{center}
\end{figure}

\section{Results}
The Score Matrix Decomposition exhibits advantageous properties compared to approximate reductionistic decomposition techniques.

\textbf{Influence of Attention Head Dimension}: Empirical results suggest a strong dependence of performance on the attention head dimension, $d_{\text{head}} \equiv \frac{d}{\#\text{Heads}}$. Specifically, experiments show that increasing $d_{\text{head}}$ from a small value (e.g., $d_{\text{head}}=16$, corresponding to a high head count $H=16$ for a fixed model dimension) to a larger value (e.g., $d_{\text{head}}=64$, corresponding to a lower head count $H=4$) yields superior performance. This observation implies that a sufficiently \textbf{large head subspace dimension} is a necessary condition for the meaningful separation and differentiation of the ${Q}$ (query) and ${K}$ (key) span spaces, which is central to the span violation decomposition process. Conversely, a \textbf{small head dimension} (corresponding to a high number of heads in our experimental setup), which results in matrices with an effectively \textbf{too small rank} for ${Q}$ and ${K}$, appears insufficient to properly demonstrate the effective decomposition.

\subsection{Training Dynamics of Reductionistic Decomposition}
The configuration  \texttt{[1100]} achieved the best performance (Loss: 5.498), outperforming the standard baseline by $0.23\%$. 

\begin{figure}[ht]
\vskip 0.1in
\begin{center}
\includegraphics[width=\columnwidth]{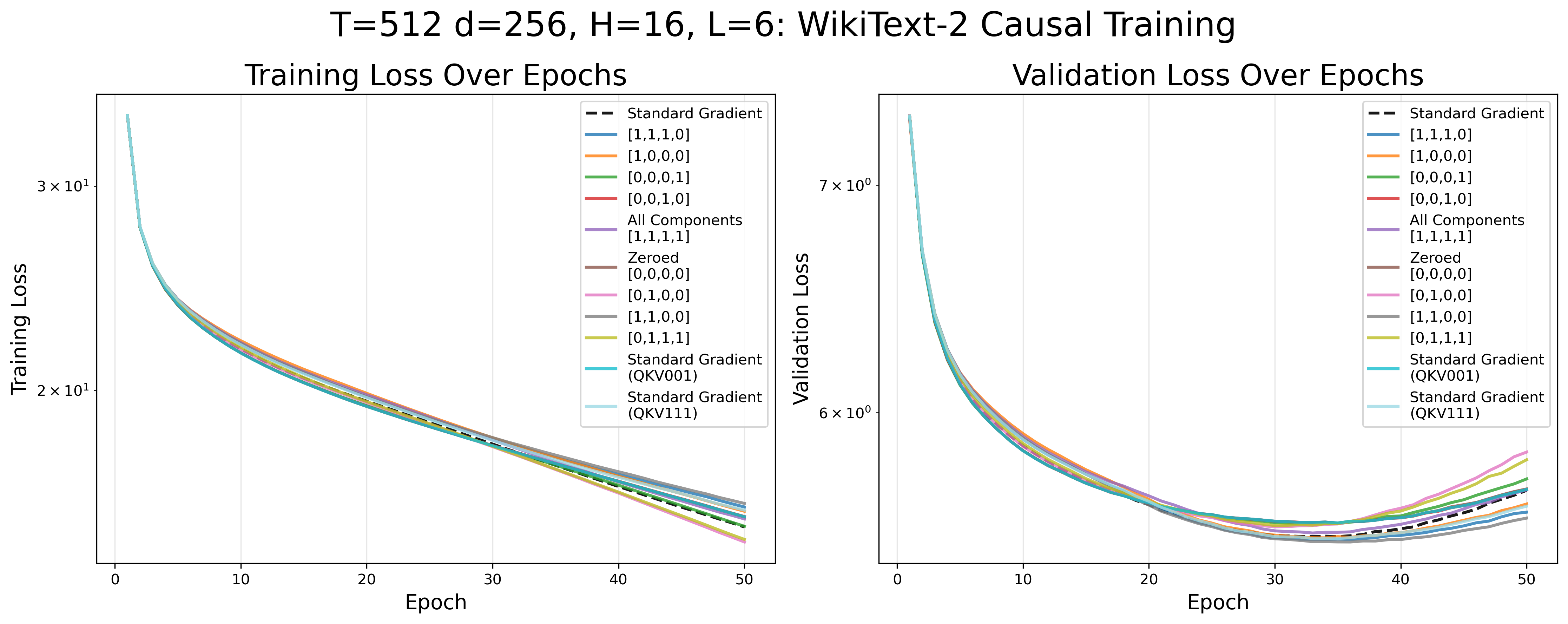}
\vskip 0.1in
\includegraphics[width=\columnwidth]{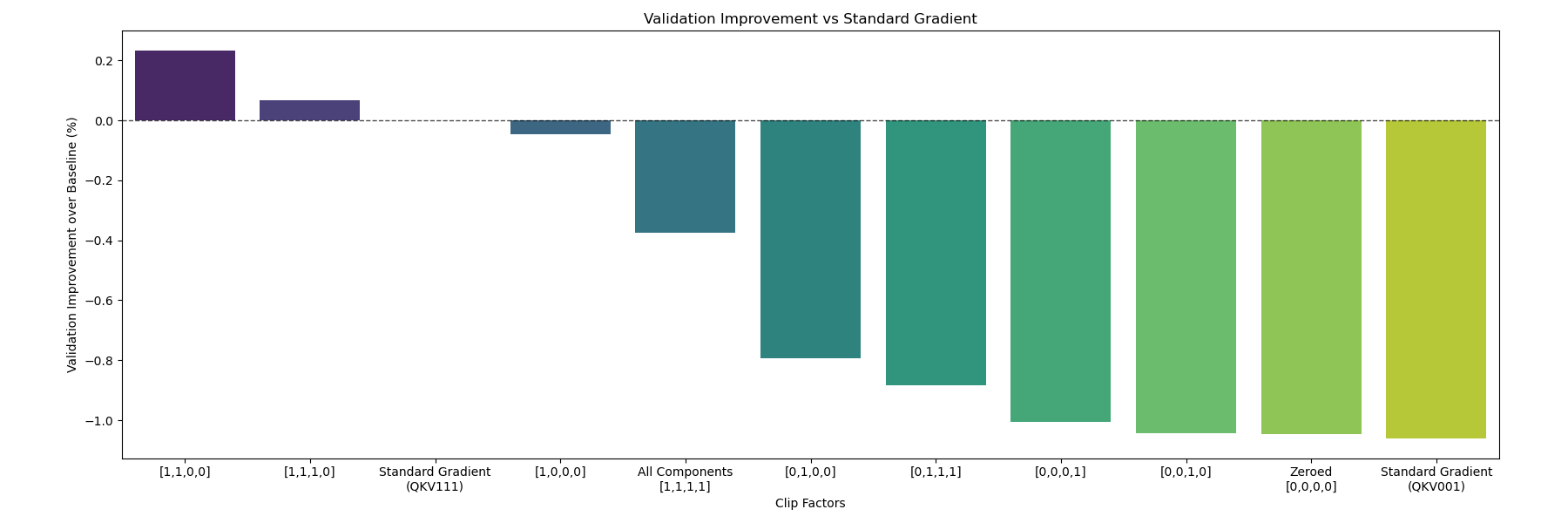}
\caption{\textbf{Reductionistic Decompositions}: Model architecture $T=512, d=256$, 16 heads($d_{Head}\equiv \frac{d}{H}=16$), 6 layers. Top: Training curves show the progression over 50 epochs on the WikiText-2 dataset causal prediction. Dashed line is the Standard Gradient baseline. Bottom: Improvements on the Standard Gradient baseline validation loss minimum.}
\label{fig:direct_causal_H16L6}
\end{center}
\end{figure}

\subsection{Training Dynamics of Score Matrix Decomposition}
Among all scaling configurations($\texttt{[$\alpha_0 \alpha_1 \alpha_2 \alpha_3$]}$),  \texttt{[1000]}, corresponding to the parallel-span–only component, achieves the strongest improvement (best-val 5.4857), outperforming the corresponding standard baseline by $0.56\%$. The next best is \texttt{[1100]}, which additionally includes the first-order span-violation term (best-val 5.5035). Both configurations outperform not only  \texttt{[1111]} but also the canonical Standard Gradient baseline. These results demonstrate that lower-order span components—especially the pure parallel span—yield the most beneficial gradient signals, whereas higher-order span-violation terms introduce noise that diminishes generalization.

\subsection{Ablation Study}

 We fix the sequence length and model dimension at $T=512$, $d=256$ and examine model behaviors under varying head and layer configurations.

\begin{table}[htbp]
\vskip 0.15in
\begin{center}
\begin{small}
\begin{sc}
\begin{tabular}{lccc}
\toprule
\#Heads & \#Layers & $d_{Head}$ & Best Enhancement\\
\midrule
1 & 1 & 256 & $\texttt{[1000]}$ 0.023$\%$ \\
4 & 6 & 64 & $\texttt{[1000]}$ 0.562$\%$ \\
16 & 6 & 16 & $\texttt{[1000]}$ 0.108$\%$ \\
\bottomrule
\end{tabular}
\end{sc}
\end{small}
\end{center}
\caption{Ablation results under $T=512$, $d=256$, batch size 128. Each entry reports the scaling factor as well as the best validation loss enhancement(higher is better) over the standard gradient over the parameters$\texttt{[$\alpha_0 \alpha_1 \alpha_2 \alpha_3$]}$ we tested.}
\label{tab:ablation_heads_layers}
\end{table}

\textbf{Minimal configuration analysis.} The single-head, single-layer configuration (H=1, L=1) shows almost no differentiation from the standard QKV baseline model. This suggests that Q and K gradients contribute minimally to training at this scale, indicating that gradient decomposition effects become significant only with increased model capacity.

\textbf{Multi-head configuration comparison.} We focus on two contrasting multi-head setups with 6 layers:
\begin{itemize}
\item \textbf{4 heads (head dim 64)}: Moderate head count with larger per-head dimension, providing substantial representational capacity per attention head while maintaining computational efficiency.
\item \textbf{16 heads (head dim 16)}: Higher head count with smaller per-head dimension, distributing attention computation across more specialized heads at the cost of reduced individual head capacity.
\end{itemize}

This comparison reveals how gradient decomposition interacts with the head dimension and number of heads trade-off, examining whether the benefits scale better with more attention heads or with richer per-head representations.

\subsubsection{ Performance Comparison}
Performance comparison of the score matrix decomposition method over standard attention gradient with varied number of heads and layers, while fixing sequence length and feature dimension is presented as:
\begin{table}[htbp]
\centering
\small
\begin{tabular}{lcccccc}
\toprule
H$\times$L & Method & \multicolumn{2}{c}{Train Loss} & \multicolumn{2}{c}{Val Loss} \\
\cmidrule(lr){3-4} \cmidrule(lr){5-6}
&  & Final & $\Delta$\% & Min & $\Delta$\% \\
\midrule
\multirow{4}{*}{1$\times$1} 
& QKV111   & 16.4481 & 0.0 & 5.6263 & 0.0 \\
& [1,1,1,1] & 16.4519 & -0.023 & 5.6262 & 0.002 \\
& [0,0,0,0] & 16.5034 & -0.336 & 5.6264 & -0.001 \\
& [1,0,0,0] & 16.4084 & 0.241 & 5.6250 & 0.023 \\
\midrule
\multirow{4}{*}{4$\times$6} 
& QKV111   & 14.6401 & 0.0 & 5.5167 & 0.0 \\
& [1,1,1,1] & 14.8002 & -1.093 & 5.5088 & 0.142 \\
& [0,0,0,0] & 15.0499 & -2.799 & 5.5524 & -0.648 \\
& [1,0,0,0] & 14.8139 & -1.186 & 5.4856 & 0.562 \\
\midrule
\multirow{4}{*}{16$\times$6} 
& QKV111   & 15.2483 & 0.0 & 5.4978 & 0.0 \\
& [1,1,1,1] & 15.1213 & 0.8328 & 5.5179 & -0.364 \\
& [0,0,0,0] & -- & -- & -- & -- \\
& [1,0,0,0] & 15.3293 & -0.531 & 5.4918 & 0.108 \\
\bottomrule
\end{tabular}
\caption{Model performance on causal prediction task with different clipping methods across $Head \times Layer$ configurations under $T=512$, $d=256$. 
The table reports final training loss, minimum validation loss, and relative percentage change ($\Delta$\%) compared to the \texttt{QKV111} baseline Standard Gradient.}
\label{tab:clipping_results}
\end{table}

\section*{Conclusion}

This work showed the canonical $O(N^2)$ Transformer can be further optimized by addressing geometric inefficiency. Experimental validation confirmed that the $0^{th}$-order span yields the most effective learning signal, though higher-order span components may become relevant at deeper abstraction levels. Acknowledging training-time overhead and evaluation on a limited dataset, these trade-offs are necessary to improve convergence quality. The most pressing directions for future research are detailed below.

\subsection*{Future Work}

Our findings open several avenues for research, primarily focusing on scaling the method and integrating it dynamically:

\begin{enumerate}
    \item \textbf{Scaling and Abstraction:} Validate the framework's full potential on massive datasets (e.g., C4, The Pile) and architectures with large head dimensions/layer depths to confirm gains at high abstraction levels.
    \item \textbf{Flexible Application:} Investigate selective decomposition (e.g., applying to upper layers) and dynamic training regimes (adjusting decomposition between early and later epochs) to optimize efficacy based on $Q, K, V$ matrix stability.
    \item \textbf{Computational Efficiency:} Develop computationally efficient, low-rank approximations or iterative methods for calculating the projection operators to reduce the training overhead.
\end{enumerate}

\section*{Acknowledgements}
A provisional patent application related to this work has been filed with the United States Patent and Trademark Office.


\bibliography{qkv}
\bibliographystyle{icml2025}

\end{document}